\documentclass[runningheads]{llncs}
\usepackage[T1]{fontenc}
\usepackage{graphicx}
\usepackage{amsmath} 

\usepackage{pifont}       
\usepackage{bbding}       
\usepackage{fontawesome}  
\usepackage{multicol}
\usepackage{multirow}
\usepackage{booktabs}
\newcommand{\cmark}{\ding{51}}
\newcommand{\xmark}{\ding{55}}

\usepackage{color}

\begin{document}
\title{Overview of the NLPCC 2026 Shared Task 1: Difficulty-Aware Multilingual and Multimodal Medical Instructional Video Understanding Evaluation}
\titlerunning{Overview of the NLPCC 2026 Shared Task 1}
\author{
    Shenxi Liu
    \inst{1}
    \and
    Kan Li
    \inst{1}
    \and
    Mingyang Zhao
    \inst{2}
    \and
    Yuhang Tian
    \inst{1}
    \and
    Bin Li
    \inst{3}\thanks{Corresponding author}
}
\authorrunning{S. Liu et al.}
\institute{
     School of Computer Science and Technology, Beijing Institute of Technology
     \email{\{liushenxi,likan,tianyuhang\}@bit.edu.cn}
     \and
     Department of Computing, The Hong Kong Polytechnic University
    \email{25019897r@connect.polyu.hk}
     \and
     Shenzhen Institute of Advanced Technology, Chinese Academy of Sciences
     \email{b.li2@siat.ac.cn}
}

\maketitle              
\begin{abstract}
Following the successful organization of the CMIVQA, MMI-VQA, and M4IVQA challenges in NLPCC 2023--2025, this year we further introduce a new shared task to advance research on difficulty-aware multilingual and multimodal medical instructional video understanding, namely the Difficulty-Aware Medical Instructional Video Question Answering (DA-MIVQA) challenge. 
Different from previous benchmarks that mainly emphasize language coverage, modality fusion, or multi-hop reasoning, DA-MIVQA explicitly introduces a difficulty-aware evaluation perspective by distinguishing questions according to the evidence required for answering, thereby better reflecting real-world medical usage scenarios. 
Specifically, the challenge categorizes evaluation questions into \emph{simple} and \emph{complex} subsets, where simple questions can often be addressed through subtitle-based textual cues, while complex questions require visual grounding, procedural understanding, and cross-modal evidence integration from medical videos. 
The DA-MIVQA challenge consists of three tracks: Difficulty-Aware Temporal Answer Grounding in Single Video (DA-TAGSV), Difficulty-Aware Video Corpus Retrieval (DA-VCR), and Difficulty-Aware Temporal Answer Grounding in Video Corpus (DA-TAGVC). 
These tracks evaluate whether participating systems can retrieve relevant instructional videos, localize answer-bearing temporal spans, and remain robust under different levels of textual, visual, and procedural complexity. 
The dataset is collected from public medical instructional channels and covers diverse scenarios such as first aid, emergency response, rehabilitation guidance, nursing practice, and general medical education, with all question-answer pairs manually verified and further annotated with difficulty labels. 
This paper presents an overview of the NLPCC 2026 shared task on DA-MIVQA, including the task motivation, dataset construction, evaluation protocol, participation overview, competition results, and representative systems. 
We believe that DA-MIVQA provides a more practical benchmark for medical instructional video question answering and will promote future advances in multimodal medical reasoning, visual grounding, and multilingual medical knowledge access. 

\end{abstract}

\section{Introduction}

\label{sec:introduction}

Recent advances in AI-assisted healthcare have shown considerable promise in a wide range of clinical and educational scenarios, including medical imaging, decision support, and procedural training \cite{scopingreview,surgicalsimulation,Shahrezaei2024}. 
Among these scenarios, medical instructional videos have become an increasingly important medium for acquiring practical skills in first aid, emergency response, nursing practice, rehabilitation, and general medical education \cite{onlinevideos,onlinevideoteaching,makingeffective}. 
Compared with textual descriptions alone, instructional videos provide intuitive, step-by-step demonstrations of medically relevant actions, tool usage, posture transitions, and temporal procedure flows, thereby offering richer evidence for medical learning and question answering\cite{Li2022LearningTL,weng2022,tovisual}. 
This practical value has motivated a series of shared tasks at NLPCC on medical instructional video question answering, including CMIVQA in 2023 \cite{nlpcc2023_cmivqa}, MMIVQA in 2024 \cite{nlpcc2024_mmivqa}, and M4IVQA in 2025 \cite{nlpcc2025_m4ivqa}. 
These benchmarks have progressively expanded the problem setting from Chinese medical instructional video question answering to multilingual and multimodal understanding, and further to multi-hop reasoning across heterogeneous evidence sources \cite{m3med}. 
In particular, prior work has made notable progress in two core directions, namely temporal answer grounding and video corpus retrieval \cite{temporal_answer_localization,video_corpus_retrieval}. 
Temporal answer grounding focuses on localizing the answer-relevant span within an untrimmed medical instructional video, while video corpus retrieval aims to identify the most relevant video from a large-scale collection given a medical question. 

Despite such progress, existing benchmarks still do not explicitly distinguish questions according to the type of evidence required for correct answering in realistic medical application scenarios \cite{multishortcoming,videochallenges}. 
In practice, some questions can be answered through straightforward textual matching or basic semantic understanding from subtitles, whereas others require direct visual grounding and contextual interpretation of procedural actions shown in the video. 

Therefore, the conventional notion of question difficulty cannot be fully characterized only by reasoning depth or the number of inference hops. 
For medical instructional video understanding, a more practical distinction lies in whether answering a question requires explicit visual evidence in addition to textual cues. 

To address this gap, NLPCC 2026 introduces the \textbf{Difficulty-Aware Medical Instructional Video Question Answering} challenge, abbreviated as \textbf{DA-MIVQA}. 
Unlike previous tasks that mainly emphasized language scope, modality coverage, or multi-hop reasoning, DA-MIVQA explicitly incorporates a difficulty-aware evaluation perspective based on the evidence structure required to answer each question. 
Specifically, DA-MIVQA categorizes questions into \emph{simple} and \emph{complex} subsets. 
Simple questions can typically be answered through subtitle-aligned textual cues, explicit single-source evidence, or basic semantic matching. 
By contrast, complex questions cannot be answered reliably from text alone and instead require visual grounding, action understanding, object-state recognition, and procedural context modeling from the video. 
This distinction makes the evaluation more aligned with realistic medical usage, where users often need not only relevant textual hints but also precise visual evidence about what to do, where to act, and when a procedure step occurs \cite{Tips,useofvideos,TimeCraft}. 

At the task level, DA-MIVQA preserves the three-track formulation established in previous NLPCC shared tasks, ensuring continuity with the existing research line \cite{nlpcc2023_cmivqa,nlpcc2024_mmivqa,nlpcc2025_m4ivqa}. 

The first track, \textbf{DA-TAGSV}, focuses on difficulty-aware temporal answer grounding in a single video, requiring systems to localize the start and end timestamps of the answer span for a given question. 
The second track, \textbf{DA-VCR}, addresses difficulty-aware video corpus retrieval, where systems must retrieve the most relevant medical instructional video from a large corpus. 
The third track, \textbf{DA-TAGVC}, combines retrieval and localization by requiring systems to first identify the relevant video and then ground the answer temporally within that video. 

By evaluating all three tracks under separate simple, complex, and mixed settings, DA-MIVQA enables a more fine-grained analysis of system robustness across different levels of procedural and cross-modal difficulty. 

The dataset of DA-MIVQA is collected from public medical instructional channels on YouTube and covers diverse medical and health-related scenarios, such as first aid, medical emergency management, rehabilitation guidance, nursing practice, and general medical education \cite{Youtube}. 
Following the annotation paradigm of the previous shared tasks, questions and temporal answers are manually verified by annotators with medical backgrounds, and each question is further labeled as simple or complex according to the type of evidence needed for answering. 
This re-annotation strategy substantially improves the benchmark's ability to reveal whether a system truly understands the visual and procedural content of medical videos, rather than relying mainly on textual matching. 

\begin{figure}[t]
    \centering
    \includegraphics[width=0.4\textwidth]{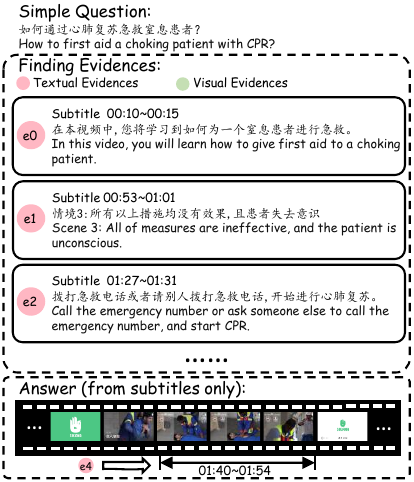} \qquad \includegraphics[width=0.4\textwidth]{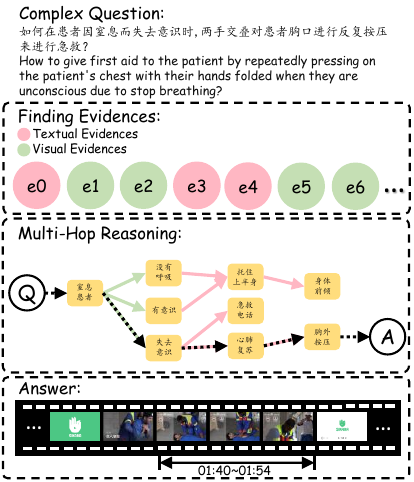}
    \caption{Dataset examples of the M4IVQA shared task.}
    \label{fig:task_evolution}
\end{figure}

\begin{table}[t]
    \centering
    \caption{A conceptual comparison of the NLPCC medical instructional video QA shared tasks.}
    \label{tab:task_comparison}
    \begin{tabular}{lcccc}
        \hline
        Task & Languages & Multi-modal & Chain-of-thought & Difficulty-aware \\
        \hline
        CMIVQA (2023) \cite{nlpcc2023_cmivqa} & Chinese & \textcolor{green}{\cmark} & \textcolor{red}{\xmark} & \textcolor{red}{\xmark} \\
        MMIVQA (2024) \cite{nlpcc2024_mmivqa} & Chinese/English & \textcolor{green}{\cmark} & \textcolor{red}{\xmark} & \textcolor{red}{\xmark} \\
        M4IVQA (2025) \cite{nlpcc2025_m4ivqa} &  Chinese/English  & \textcolor{green}{\cmark} & \textcolor{green}{\cmark} &\textcolor{red}{\xmark} \\
        DA-MIVQA (2026) & Chinese/English  & \textcolor{green}{\cmark} & \textcolor{green}{\cmark} & \textcolor{green}{\cmark} \\
        \hline
    \end{tabular}
\end{table}

From an application perspective, DA-MIVQA is intended to support the development of more reliable medical video question answering systems for education, skill acquisition, emergency guidance, and multilingual knowledge access. 

More importantly, it provides a benchmark that can distinguish systems that merely match subtitles from those that can genuinely integrate visual, textual, and procedural evidence in medically grounded scenarios. 

In this paper, we present an overview of the NLPCC 2026 shared task on DA-MIVQA, including its motivation, task definition, dataset construction, evaluation protocol, participation summary, and representative systems. 
We hope that this challenge will encourage future research on difficulty-aware medical video understanding and contribute to more practical multimodal and multilingual medical AI systems.

\section{Task Introduction}
\label{sec:task_intro}

The DA-MIVQA challenge aims to promote the development of medical instructional video question answering systems under a difficulty-aware evaluation setting. 

Different from previous shared tasks that primarily emphasized multilingual understanding, multimodal fusion, or multi-hop reasoning \cite{nlpcc2023_cmivqa,nlpcc2024_mmivqa,nlpcc2025_m4ivqa,m3med}, DA-MIVQA explicitly evaluates whether systems remain effective across questions with different evidence requirements \cite{assessingmodality,tvqa}. 
In this challenge, all questions are categorized into \emph{simple} and \emph{complex} subsets according to the type of evidence required for answering, so that model performance can be analyzed not only in terms of overall effectiveness but also in terms of robustness under different levels of procedural and cross-modal complexity. 
Simple questions are mainly answerable through subtitle-aligned textual matching or basic semantic understanding, whereas complex questions require visual grounding, procedural interpretation, and the integration of textual and visual evidence from medical instructional videos. 
Following the design paradigm of the previous NLPCC medical video shared tasks, DA-MIVQA is organized into three tracks, namely Difficulty-Aware Temporal Answer Grounding in Single Video, Difficulty-Aware Video Corpus Retrieval, and Difficulty-Aware Temporal Answer Grounding in Video Corpus. 

\subsection{Definition of Each Track}
\label{subsec:track_definition}

The proposed challenge contains three complementary tracks that cover the core capabilities required in medical instructional video understanding, including temporal localization, corpus retrieval, and their joint modeling. 

\begin{figure}[t]
    \centering
    \includegraphics[width=0.9\textwidth,trim=0cm 0.15cm 0cm 0cm,clip]{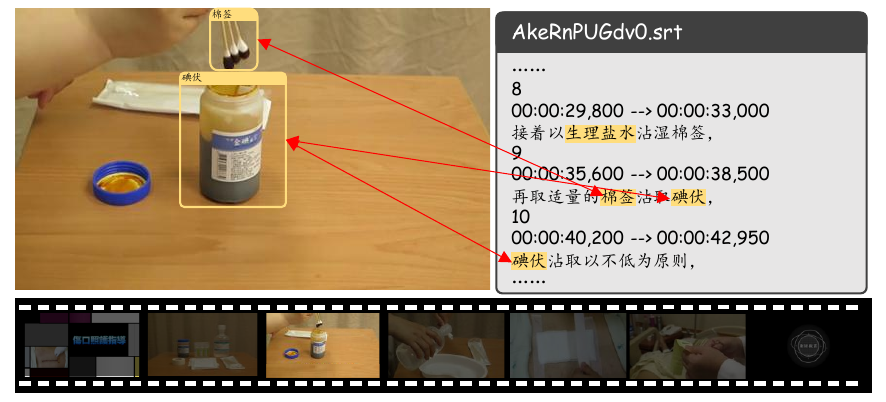}
    \caption{The multi-modal features of complex questions in DA-MIVQA.}

    \label{fig:da_mivqa_tracks}
\end{figure}

\paragraph{Track 1: Difficulty-Aware Temporal Answer Grounding in Single Video (DA-TAGSV).}
Given a medical or health-related question and a single untrimmed medical instructional video, this track aims to locate the temporal answer span within the video, that is, the start and end timestamps of the segment that best answers the input question. 
The purpose of this track is to evaluate whether a system can perform fine-grained temporal localization under a difficulty-aware setting, especially when the answer may depend either on explicit subtitle cues or on visually grounded procedural evidence. 

For simple questions, the target span may often be inferred from subtitle-level textual evidence, while for complex questions, the system is expected to localize the relevant segment by understanding medical actions, object states, and procedural transitions shown in the video. 

\begin{figure}[t]
    \centering
    \includegraphics[width=0.9\textwidth]{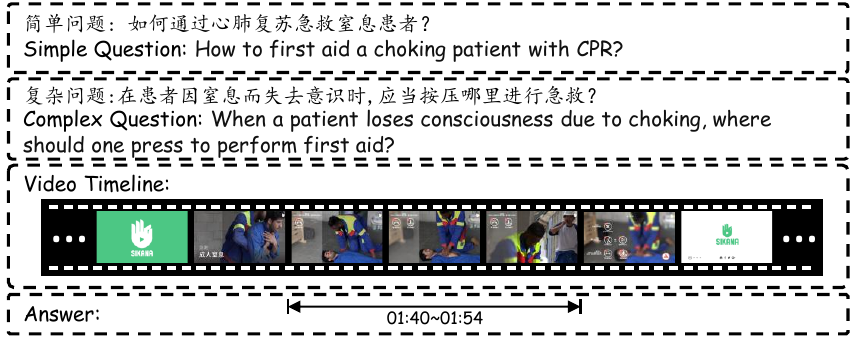}
    \caption{Illustration of Track 1: Difficulty-Aware Temporal Answer Grounding in Single Video (DA-TAGSV).}
    \label{fig:track1_da_tagsv}
\end{figure}

\paragraph{Track 2: Difficulty-Aware Video Corpus Retrieval (DA-VCR).}
Given a medical or health-related question and a large collection of untrimmed medical instructional videos, this track aims to retrieve the most relevant video in the corpus for the input question. 
This track evaluates the capability of a system to perform semantic matching between a question and candidate videos under multilingual and multimodal conditions, while further examining whether retrieval performance remains stable across simple and complex questions. 

For simple questions, retrieval may rely more on subtitle-level lexical or semantic overlap, whereas complex questions often require systems to infer which video contains the most relevant visually grounded procedure or action sequence. 

\begin{figure}[t]
    \centering
    \includegraphics[width=0.9\textwidth]{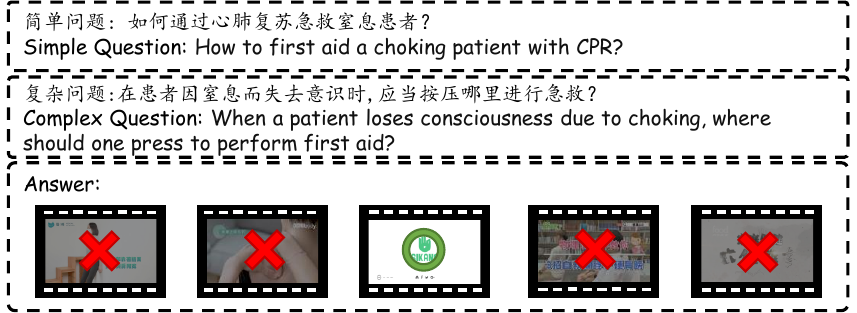}
    \caption{Illustration of Track 2: Difficulty-Aware Video Corpus Retrieval (DA-VCR).}
    \label{fig:track2_da_vcr}
\end{figure}

\paragraph{Track 3: Difficulty-Aware Temporal Answer Grounding in Video Corpus (DA-TAGVC).}
Given a medical or health-related question and a large collection of untrimmed medical instructional videos, this track first requires the system to retrieve the most relevant video from the corpus and then to locate the corresponding temporal answer span within that video. 
This track combines the challenges of the first two tracks and therefore evaluates the holistic ability of a system in video retrieval, answer localization, multimodal understanding, and difficulty-aware reasoning. 

Compared with the first two tracks, this setting is more demanding because retrieval errors and localization errors may accumulate, especially for complex questions that require both accurate video selection and visually grounded temporal reasoning. 

\begin{figure}[t]
    \centering
    \includegraphics[width=0.9\textwidth]{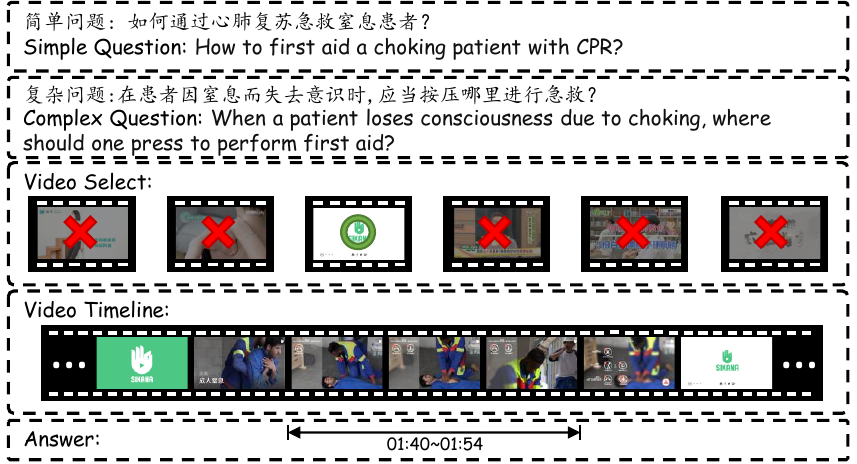}
    \caption{Illustration of Track 3: Difficulty-Aware Temporal Answer Grounding in Video Corpus (DA-TAGVC).}
    \label{fig:track3_da_tagvc}
\end{figure}

\subsection{Evaluation Metrics}
\label{subsec:eval_metrics}

Following the evaluation style of previous NLPCC medical instructional video shared tasks, DA-MIVQA adopts task-specific metrics for each track while additionally reporting results on simple-only, complex-only, and mixed subsets \cite{nlpcc2023_cmivqa,nlpcc2024_mmivqa,nlpcc2025_m4ivqa}. 
This evaluation protocol allows the challenge to measure not only the overall performance of participating systems but also their robustness under different levels of evidence complexity. 

\paragraph{Metrics for Track 1: DA-TAGSV.}
For the single-video temporal grounding track, the evaluation is based on temporal overlap between predicted answer spans and gold answer spans. 
We adopt the standard Intersection over Union (IoU) metric and the mean Intersection over Union (mIoU), where mIoU denotes the average IoU over all test samples \cite{temporal_answer_localization}. 
In addition, we report the metric \emph{R@1, IoU = $\mu$}, where $\mu \in \{0.3, 0.5, 0.7\}$, to measure the proportion of test samples for which the top-ranked predicted span achieves an IoU no smaller than the corresponding threshold. 
The main ranking metric of this track is mIoU, while the other metrics are provided for detailed analysis. 

\begin{equation}
\mathrm{IoU}=\frac{A \cap B}{A \cup B}
\label{eq:iou_da_tagsv}
\end{equation}

\begin{equation}
\mathrm{mIoU}=\frac{1}{N}\sum_{i=1}^{N}\mathrm{IoU}_i
\label{eq:miou_da_tagsv}
\end{equation}

\paragraph{Metrics for Track 2: DA-VCR.}
For the video corpus retrieval track, we adopt retrieval-oriented metrics to evaluate whether the relevant video is ranked highly in the returned list. 
Specifically, we use recall-based metrics \emph{R@n} with $n=1,10,100$ to measure retrieval performance at different cut-off depths. 
We also adopt the Mean Reciprocal Rank (MRR) metric to evaluate the ranking position of the target video in the predicted list \cite{video_corpus_retrieval}. 
To provide a unified score for ranking participating systems, we further report the \emph{Overall} score, which is calculated as the average of R@1, R@10, R@100, and MRR. 
The main ranking metric of this track is the Overall score. 

\begin{equation}
\mathrm{MRR}=\frac{1}{|V|}\sum_{i=1}^{|V|}\frac{1}{\mathrm{Rank}_i}
\label{eq:mrr_da_vcr}
\end{equation}

\begin{equation}
\mathrm{Overall}=\frac{1}{|M|}\sum_{i=1}^{|M|}\mathrm{Value}_i
\label{eq:overall_da_vcr}
\end{equation}

\paragraph{Metrics for Track 3: DA-TAGVC.}
For the corpus-level temporal answer grounding track, we combine retrieval-based metrics and localization-based metrics to jointly evaluate whether a system can retrieve the correct video and accurately localize the answer span within it. 
We retain the temporal localization metric IoU from Track 1 and the retrieval-oriented metrics from Track 2 for detailed analysis. 
In addition, we report \emph{R@1|mIoU}, \emph{R@10|mIoU}, and \emph{R@100|mIoU}, which reflect the average localization quality under different retrieval depths. 
For final ranking, we adopt the \emph{Average} score, which is computed as the mean of R@1|mIoU, R@10|mIoU, and R@100|mIoU. 
The main ranking metric of this track is the Average score. 

\begin{equation}
\mathrm{Average}=\frac{1}{|M'|}\sum_{i=1}^{|M'|}\mathrm{Value}'_i
\label{eq:average_da_tagvc}
\end{equation}

\subsection{Dataset}
\label{subsec:dataset_da_mivqa}

The dataset of DA-MIVQA is collected from public medical instructional channels on YouTube and covers a wide range of medical and health-related scenarios \cite{EffectiveEdu}. 
These scenarios include first aid, medical emergency management, rehabilitation guidance, nursing practice, and general medical education, ensuring the practical relevance and topical diversity of the benchmark. 
Following the annotation paradigm of previous shared tasks, the questions and corresponding temporal answers are manually verified by annotators with medical backgrounds to ensure annotation quality and medical validity \cite{Gupta2023}. 
Each video may contain multiple question-answer pairs, and each question is associated with a unique temporal answer span marked by start and end timestamps. 
On top of the original annotations, DA-MIVQA further introduces difficulty labels, namely \emph{simple} and \emph{complex}, to reflect the type of evidence required for answering each question. 
This re-annotation strategy makes it possible to distinguish text-dominant matching questions from genuinely multimodal procedural understanding questions in medical instructional videos\cite{medcraft}. 
In particular, simple questions can usually be answered from subtitle-aligned textual cues or other explicit single-source evidence, while complex questions require visual grounding together with subtitle and temporal procedural context. 
Therefore, the dataset is designed not only for evaluating answer correctness but also for analyzing whether a model truly captures medically relevant actions, object states, tools, posture changes, and procedure transitions shown in videos. 

In terms of scale, the dataset is divided into training, validation, and test subsets, and each subset contains both Chinese and English questions with simple and complex labels. 
Compared with previous shared tasks, the present dataset is expanded through difficulty-aware re-annotation and an increased number of question-answer pairs, thus providing a more fine-grained benchmark for multilingual and multimodal medical instructional video understanding. 

\begin{table}[htbp]
\centering
\caption{Statistics of Sample Counts for the Three Tracks}
\label{tab:track_statistics}
\begin{tabular}{lllcccc}
\toprule
Track & Language & Complexity & Training Set & Validation Set & Test Set & Total \\
\midrule
\multirow{4}{*}{Track 1}
	&	 Chinese 	&	 Simple  	&	2101	&	362	&	351	&	2814	  \\
	&	         	&	 Complex 	&	1842	&	285	&	339	&	2466	  \\
	&	 English 	&	 Simple  	&	1273	&	199	&	202	&	1674	  \\
	&	         	&	 Complex 	&	1341	&	223	&	219	&	1783	  \\
\addlinespace													\midrule
\multirow{4}{*}{Track 2}													
	&	 Chinese 	&	 Simple  	&	2101	&	362	&	394	&	2857	  \\
	&	         	&	 Complex 	&	1842	&	285	&	280	&	2407	  \\
	&	 English 	&	 Simple  	&	1273	&	199	&	210	&	1682	  \\
	&	         	&	 Complex 	&	1341	&	223	&	234	&	1798	  \\
\addlinespace													\midrule
\multirow{4}{*}{Track 3}													
	&	 Chinese 	&	 Simple  	&	2101	&	362	&	394	&	2857	  \\
	&	         	&	 Complex 	&	1842	&	285	&	280	&	2407	  \\
	&	 English 	&	 Simple  	&	1273	&	199	&	210	&	1682	  \\
	&	         	&	 Complex 	&	1341	&	223	&	234	&	1798	  \\

\bottomrule
\end{tabular}
\end{table}

\section{Baseline}
\label{sec:baseline}

To provide meaningful reference points for the proposed DA-MIVQA evaluation, we design two complementary baseline settings.

The first baseline adopts a high-performance large language model as a text-only oracle, where only subtitle transcripts are provided as input.
This setting is intended to estimate the upper-bound performance that can be achieved when the system relies solely on textual cues without access to visual content.

The second baseline is based on our multimodal knowledge mining and encoder--decoder framework, which incorporates both textual and visual information from medical instructional videos.
This setting provides a practical multimodal reference for evaluating how much additional benefit can be obtained by modeling visual evidence, temporal context, and cross-modal interactions.

Together, these two baselines allow us to examine the performance gap between text-only reasoning and multimodal video understanding under different evidence requirements.

\subsection{Text-only Oracle Baseline}

For the text-only oracle baseline, we use a frozen Qwen3.5-9B\cite{qwen35blog} model as the backbone large language model.
Instead of updating the parameters of the backbone model, we adapt the system to DA-MIVQA by tuning task-specific system prompts.
During training, the timestamped SRT subtitles from the training set are used as the user-side input prompts.
Each training instance consists of a question, the corresponding timestamped subtitle sequence, and the expected task-specific output.

For DA-TAGSV, the model is trained to predict the temporal span in the given video that best supports the answer.
For DA-VCR, the model is trained to identify the most relevant video from the candidate corpus based on subtitle-level textual evidence.
For DA-TAGVC, the model is trained to first infer the relevant video and then predict the answer-supporting temporal span within that video.

As a result, we obtain three task-specific text-only oracle models, corresponding to DA-TAGSV, DA-VCR, and DA-TAGVC, respectively.
This baseline is not intended to represent a deployable multimodal solution, but rather to quantify how far subtitle-only reasoning can go under different evidence requirements.

\subsection{Multi-modal Knowledge Mining and Encoder--Decoder Baseline}

\begin{figure}[t]
    \centering
    \includegraphics[width=0.9\textwidth]{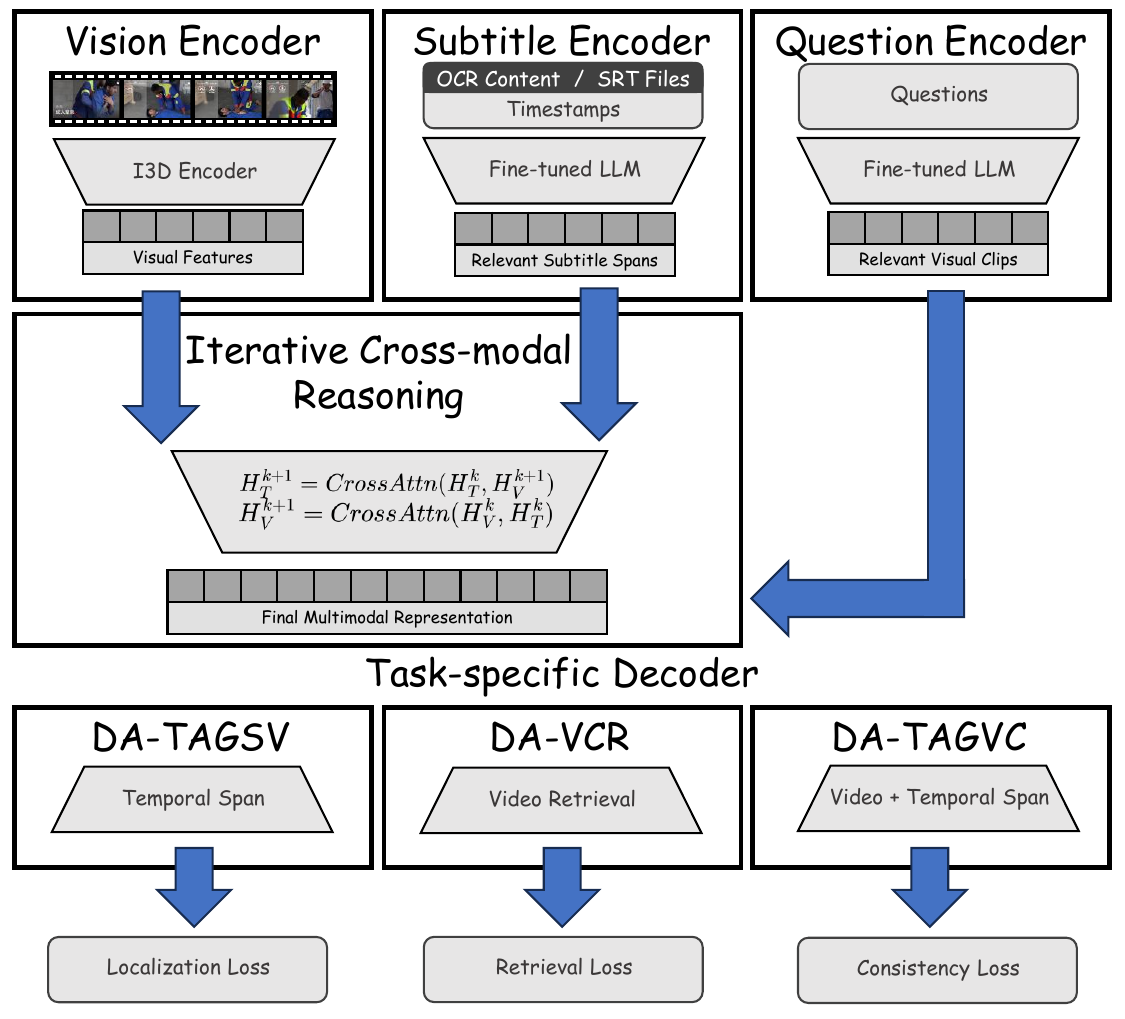}
    \caption{Overview of multi-modal knowledge mining and encoder--decoder baseline method.}

\end{figure}

For the multi-modal baseline, we adopt an iterative cross-modal reasoning strategy as the main improvement direction.
This choice is motivated by the evidence-aware nature of DA-MIVQA, where some questions can be answered from subtitles alone, whereas others require visual grounding and temporal refinement.

The multi-modal baseline is conceptually inspired by prior work on MutualSL (visual answer localization with cross-modal mutual knowledge transfer) \cite{mutualsl}, but it is not a direct reproduction of the original model.
Instead, we extend the underlying idea to the DA-MIVQA setting by tailoring it to the evidence-aware characteristics of our evaluation task.
Specifically, we adapt cross-modal mutual knowledge transfer into an iterative evidence mining and encoder--decoder framework that jointly models subtitles, visual clips, questions, and temporal answer spans.

Compared with explicit graph-based reasoning, iterative cross-modal attention provides a lightweight and end-to-end trainable mechanism for modeling multi-hop interactions among questions, subtitles, and video clips.
Specifically, we build the baseline as a multimodal knowledge mining and encoder--decoder framework.

The framework first extracts timestamped subtitle representations, question representations, and uniformly sampled visual clip representations from each medical instructional video.
The subtitle encoder maps each SRT sentence into a contextual textual representation while preserving its start and end timestamps.
The visual encoder maps each sampled clip into a visual representation that captures actions, objects, scenes, and procedural cues.
After initial encoding, the model performs multi-modal knowledge mining by retrieving subtitle spans and video clips that are semantically relevant to the question.
The retrieved textual and visual candidates are then passed into an iterative cross-modal reasoning block.
At each reasoning hop, the question-enhanced subtitle representation attends to the visual clips, and the visual representation then attends back to the subtitle sequence.
Formally, given textual states $H_T^k$ and visual states $H_V^k$ at the $k$-th hop, the reasoning process is defined as follows:

\begin{align}
H_V^{k+1} = \mathrm{CrossAttn}(H_V^k, H_T^k),
\quad
H_T^{k+1} = \mathrm{CrossAttn}(H_T^k, H_V^{k+1}).
\end{align}

This iterative design allows the model to follow reasoning paths such as question-to-subtitle, subtitle-to-video, and video-to-temporal-context.
The final multimodal representation is fed into a decoder that produces task-specific predictions for the three DA-MIVQA subtasks.
For DA-TAGSV, the decoder predicts the start and end timestamps of the answer-supporting segment in the given video.
For DA-VCR, the decoder scores candidate videos according to their multimodal relevance to the question and returns the most relevant video.
For DA-TAGVC, the decoder first identifies the relevant video from the corpus and then predicts the answer-supporting temporal span within that video.

To stabilize training, we optimize the framework with a combination of localization loss, retrieval loss, and cross-modal consistency loss.
The localization loss supervises the predicted temporal boundaries, while the retrieval loss supervises the ranking of candidate videos.
The cross-modal consistency loss encourages the textual and visual branches to focus on temporally aligned evidence when producing their predictions.

Overall, this baseline provides a practical multimodal reference for DA-MIVQA by combining subtitle cues, visual demonstrations, and temporal evidence within a unified encoder--decoder architecture.
Unlike the text-only oracle baseline, this model is expected to better handle complex questions that require evidence about what action is performed, where it occurs, and when the relevant procedure step appears.

\section{Evaluation Results}
\label{sec:evaluation_results}

In this section, we summarize the participation status and evaluation results of the NLPCC 2026 shared task on Difficulty-Aware Medical Instructional Video Question Answering (DA-MIVQA). 

Following the reporting style of previous NLPCC shared task overview papers, we present the overall participation overview, the official rankings of each track, and brief comparative observations across different difficulty settings \cite{nlpcc2023_cmivqa,nlpcc2024_mmivqa,nlpcc2025_m4ivqa}. 

\subsection{Participation Overview}
\label{subsec:participation_overview}

The DA-MIVQA challenge attracted teams from universities, research institutes, and industry participants interested in multilingual and multimodal medical video understanding. 

For NLPCC2025 Shared Task 4, a total of 22 teams registered for Track 1, 14 teams for Track 2, and 16 teams for Track 3. During the test phase, 17, 10, and 11 teams submitted valid results, respectively, as shown in Table~\ref{tab:participation_overview_da_mivqa}. In the end, \textbf{HoMaMaOvO}, \textbf{WuKong}, and \textbf{Chiikawa} achieved the SOTA for Tracks 1, 2, and 3.
\begin{table}[htbp]
    \centering
    \caption{Overview of participation statistics in the DA-MIVQA challenge.}
    \label{tab:participation_overview_da_mivqa}
    \begin{tabular}{lcc}
        \hline
        Track & Registered teams & Valid submissions  \\
        \hline
        DA-TAGSV & 22 & 17  \\
        DA-VCR & 14 & 10 \\
        DA-TAGVC & 16 & 11 \\
        \hline
    \end{tabular}
\end{table}

In general, the participation distribution across the three tracks can reflect the relative technical difficulty of each task setting, especially because DA-TAGVC requires systems to jointly solve retrieval and temporal grounding under difficulty-aware evaluation. 

\subsection{Results of Track 1: DA-TAGSV}
\label{subsec:results_track1}

Track 1, Difficulty-Aware Temporal Answer Grounding in Single Video, evaluates whether a system can accurately localize the temporal answer span in an untrimmed medical instructional video for a given question. 

The official ranking of this track is based on the \emph{mIoU} score, while \emph{R@1, IoU = 0.3}, \emph{R@1, IoU = 0.5}, and \emph{R@1, IoU = 0.7} are reported as complementary metrics for detailed analysis. 

Table~\ref{tab:track1_results_da_mivqa} presents the final ranking results for Track 1. 

\begin{table}[htbp]
    \centering
    \caption{Results of Track 1: Difficulty-Aware Temporal Answer Grounding in Single Video (DA-TAGSV).}
    \label{tab:track1_results_da_mivqa}
    \begin{tabular*}{\linewidth}{@{\extracolsep{\fill}}llcccc}
    \hline
    Rank & Team ID & R@1,IoU=0.3 & R@1,IoU=0.5 & R@1,IoU=0.7 & mIoU \\
    \hline
1	&	Amazon Inc.	&	\textbf{0.5512} 	&	0.4007 	&	0.2217 	&	\textbf{0.3912} 	\\
2	&	Karamay	&	0.5334 	&	\textbf{0.4110} 	&	\textbf{0.2269} 	&	0.3904 	\\
3	&	Ouc\_AI \cite{zhang2024improving}	&	0.5101 	&	0.3579 	&	0.2146 	&	0.3608 	\\
4	&	Dvoe protiv vetra	&	0.4946 	&	0.3688 	&	0.2183 	&	0.3606 	\\
5	&	SETAG\cite{zhou2023improving}	&	0.4116 	&	0.3363 	&	0.1790 	&	0.3089 	\\
6	&	HIIT	&	0.4191 	&	0.3287 	&	0.1760 	&	0.3079 	\\
7	&	MM-Baseline \cite{mutualsl}	&	0.4265 	&	0.3211 	&	0.1729 	&	0.3068 	\\
8	&	Text-Only Baseline \cite{qwen35blog} 	&	0.4206 	&	0.3127 	&	0.1441 	&	0.2925 	\\
9	&	BFSU	&	0.4147 	&	0.3043 	&	0.1153 	&	0.2782 	\\
10	&	Random Pick Method \cite{li2023overview}	&	0.0774 	&	0.0818 	&	0.0403 	&	0.0665 	\\
    \hline
    \end{tabular*}
\end{table}

Under the difficulty-aware setting, Track 1 is expected to reveal a clear performance gap between simple and complex questions, because the latter require more reliable visual grounding and procedural understanding beyond subtitle-based matching. 

Compared with simple questions, complex questions in this track typically place higher demands on identifying medically relevant actions, body positions, tool usage, and temporal transitions in the video. 

\subsection{Results of Track 2: DA-VCR}
\label{subsec:results_track2}

Track 2, Difficulty-Aware Video Corpus Retrieval, evaluates whether a system can retrieve the most relevant medical instructional video from a large corpus for an input question \cite{video_corpus_retrieval}. 

The official ranking of this track is based on the \emph{Overall} score, which summarizes the retrieval quality measured by \emph{R@1}, \emph{R@10}, \emph{R@100}, and \emph{MRR}. 

Table~\ref{tab:track2_results_da_mivqa} shows the final ranking results for Track 2. 

\begin{table}[htbp]
    \centering
    \caption{Results of Track 2: Difficulty-Aware Video Corpus Retrieval (DA-VCR).}
    \label{tab:track2_results_da_mivqa}
    \begin{tabular*}{\linewidth}{@{\extracolsep{\fill}}llccccc}
    \hline
    Rank & Team ID & R@1 & R@10 & R@100 & MRR & Overall \\
    \hline
1	&	Team\_Wukong	&	\textbf{0.3751} 	&	0.4150 	&	0.5189 	&	\textbf{0.3931} 	&	\textbf{0.4255} 	\\
2	&	Chaldeas	&	0.3559 	&	\textbf{0.4160} 	&	\textbf{0.5492} 	&	0.3537 	&	0.4187 	\\
3	&	DIMA\cite{dima2025}	&	0.3268 	&	0.4140 	&	0.5260 	&	0.3325 	&	0.3999 	\\
4	&	sun \cite{yu2024mqua}	&	0.3086 	&	0.3721 	&	0.4442 	&	0.3000 	&	0.3562 	\\
5	&	HDU\_Team2	&	0.3139 	&	0.3647 	&	0.4447 	&	0.2982 	&	0.3554 	\\
6	&	DSG-1 \cite{lei2023two}	&	0.3192 	&	0.3572 	&	0.4452 	&	0.2964 	&	0.3545 	\\
7	&	MM-Baseline \cite{mutualsl}	&	0.1570 	&	0.1953 	&	0.2046 	&	0.1546 	&	0.1779 	\\
8	&	AC Automation	&	0.1518 	&	0.1712 	&	0.1704 	&	0.1461 	&	0.1599 	\\
9	&	Text-Only Baseline \cite{qwen35blog} 	&	0.1465 	&	0.1470 	&	0.1361 	&	0.1375 	&	0.1418 	\\
10	&	Random Pick Method \cite{li2023overview}	&	0.0361 	&	0.0822 	&	0.0962 	&	0.0636 	&	0.0695 	\\
    \hline
    \end{tabular*}
\end{table}

Compared with Track 1, Track 2 places greater emphasis on identifying question-video relevance at the corpus level, and therefore tests the semantic matching ability of systems under multilingual and multimodal conditions. 

Under the simple setting, retrieval performance may benefit more from lexical overlap and subtitle-level semantic clues, whereas the complex setting is expected to require stronger modeling of visually grounded procedures and action semantics. 

\subsection{Results of Track 3: DA-TAGVC}
\label{subsec:results_track3}

Track 3, Difficulty-Aware Temporal Answer Grounding in Video Corpus, is the most comprehensive setting in DA-MIVQA because it requires systems to retrieve the relevant video and localize the answer span within that video at the same time. 

The official ranking of this track is based on the \emph{Average} score, which is computed from \emph{R@1|mIoU}, \emph{R@10|mIoU}, and \emph{R@100|mIoU}. 

Table~\ref{tab:track3_results_da_mivqa} provides the final ranking results for Track 3. 

\begin{table}[htbp]
    \centering
    \caption{Results of Track 3: Difficulty-Aware Temporal Answer Grounding in Video Corpus (DA-TAGVC).}
    \label{tab:track3_results_da_mivqa}
    \begin{tabular*}{\linewidth}{@{\extracolsep{\fill}}llcccc}
    \hline
    Rank & Team ID & R@10,mIoU & R@100,mIoU & R@1,mIoU & Average \\
    \hline
1	&	BIGC	&	\textbf{0.1646} 	&	\textbf{0.2879} 	&	\textbf{0.3660} 	&	\textbf{0.2728} 	\\
2	&	UWM	&	0.1415 	&	0.2727 	&	0.3338 	&	0.2493 	\\
3	&	IIEleven\cite{ma2024multilingual}	&	0.1238 	&	0.2513 	&	0.3566 	&	0.2439 	\\
4	&	UESC	&	0.1364 	&	0.2551 	&	0.3359 	&	0.2425 	\\
5	&	MedEcho\cite{medecho2025}	&	0.1490 	&	0.2588 	&	0.3152 	&	0.2410 	\\
6	&	Nsddd\cite{cheng2023unified}	&	0.1436 	&	0.2129 	&	0.3235 	&	0.2267 	\\
7	&	DesiWen	&	0.1232 	&	0.2351 	&	0.3159 	&	0.2248 	\\
8	&	MM-Baseline \cite{mutualsl}	&	0.1028 	&	0.2572 	&	0.3082 	&	0.2228 	\\
9	&	Text-Only Baseline \cite{qwen35blog} 	&	0.0885 	&	0.2415 	&	0.2853 	&	0.2051 	\\
10	&	Random Pick Method \cite{li2023overview}	&	0.0454 	&	0.0872 	&	0.0673 	&	0.0666 	\\
    \hline
    \end{tabular*}
\end{table}

Because retrieval and localization errors may accumulate in this track, its performance is expected to be lower than that of the first two tracks, especially on complex questions requiring both correct video selection and accurate visual-temporal grounding. 

Therefore, the results of this track are particularly useful for examining the end-to-end capability of systems in realistic medical instructional video question answering scenarios. 

\subsection{Comparative Observations}
\label{subsec:comparative_observations}

Across all three tracks, the difficulty-aware setting makes it possible to compare system behavior on text-dominant and visually grounded questions in a more fine-grained manner. 
The expected comparison between \emph{Simple Only} and \emph{Complex Only} results can help reveal whether a model mainly relies on subtitle matching or can truly integrate textual, visual, and procedural evidence. 
In particular, if a system maintains relatively stable performance across the two difficulty subsets, it may indicate stronger robustness in handling medically grounded multimodal reasoning. 
By contrast, a large performance drop from simple to complex questions would suggest that current methods still have limitations in visual grounding, action understanding, and temporal procedural modeling for medical videos.

\section{Conclusion}
\label{sec:conclusion}

In this paper, we presented an overview of the NLPCC 2026 shared task on Difficulty-Aware Medical Instructional Video Question Answering (DA-MIVQA), which extends the research line of CMIVQA, MMIVQA, and M4IVQA toward a more fine-grained and practically grounded evaluation setting. 

Different from previous benchmarks that mainly focused on language coverage, modality expansion, or multi-hop reasoning, DA-MIVQA explicitly introduced a difficulty-aware perspective by distinguishing questions according to the type of evidence required for answering. 
Under this setting, simple questions can often be answered from subtitle-aligned textual cues, whereas complex questions require stronger visual grounding, procedural understanding, and cross-modal evidence integration from medical instructional videos. 

To support a comprehensive evaluation, the shared task was organized into three tracks, namely Difficulty-Aware Temporal Answer Grounding in Single Video, Difficulty-Aware Video Corpus Retrieval, and Difficulty-Aware Temporal Answer Grounding in Video Corpus. 
These tracks jointly evaluate temporal localization, corpus retrieval, and end-to-end retrieval-grounding ability in multilingual and multimodal medical instructional video understanding scenarios. 
In addition, the dataset was constructed from public medical instructional video sources and further enriched with difficulty annotations, making it possible to analyze whether systems truly understand medically relevant actions, object states, procedural transitions, and temporal evidence in videos. 
The evaluation framework of DA-MIVQA provides a more practical benchmark for distinguishing systems that mainly rely on subtitle matching from those that can genuinely integrate textual, visual, and procedural information in medically grounded settings. 

We expect that the DA-MIVQA challenge will encourage future research on difficulty-aware medical video understanding and promote the development of more robust multimodal and multilingual medical question answering systems. 
More broadly, this benchmark may contribute to the advancement of practical medical AI applications in education, emergency guidance, rehabilitation training, and cross-lingual knowledge access. 

In future work, richer baseline systems, stronger multi-modal reasoning models, and more fine-grained diagnostic analyses can be incorporated to further reveal the remaining challenges in medical instructional video question answering. 
Overall, DA-MIVQA represents a meaningful step toward more realistic, interpretable, and difficulty-aware evaluation for multilingual and multi-modal medical instructional video understanding.

\section*{Acknowledgement}
This work was supported by National Natural Science Foundation of China (Nos. 4222037 and L181010), and Sanming Project of Medicine in Shenzhen (No. SZZYSM202311002).

\bibliographystyle{unsrt}
\bibliography{ref}

@String{Computing = "Computing" }

@String{Computer = "{IEEE} Computer" }

@String{Springer = "Springer-Verlag" }

@InProceedings{nlpcc2023_cmivqa,
author="Li, Bin
and Weng, Yixuan
and Guo, Hu
and Sun, Bin
and Li, Shutao
and Luo, Yuhao
and Qi, Mengyao
and Liu, Xufei
and Han, Yuwei
and Liang, Haiwen
and Gao, Shuting
and Chen, Chen",
editor="Liu, Fei
and Duan, Nan
and Xu, Qingting
and Hong, Yu",
title="Overview of the NLPCC 2023 Shared Task: Chinese Medical Instructional Video Question Answering",
booktitle="Natural Language Processing and Chinese Computing",
year="2023",
publisher="Springer Nature Switzerland",
address="Cham",
pages="233--242",
isbn="978-3-031-44699-3"
}

@InProceedings{nlpcc2024_mmivqa,
author="Li, Bin
and Weng, Yixuan
and Song, Qiya
and Liang, Lianhui
and Min, Xianwen
and Zhou, Shoujun",
editor="Wong, Derek F.
and Wei, Zhongyu
and Yang, Muyun",
title="Overview of the NLPCC 2024 Shared Task 7: Multi-lingual Medical Instructional Video Question Answering",
booktitle="Natural Language Processing and Chinese Computing",
year="2025",
publisher="Springer Nature Singapore",
address="Singapore",
pages="429--439",
isbn="978-981-97-9443-0"
}

@InProceedings{nlpcc2025_m4ivqa,
author="Li, Bin
and Liu, Shenxi
and Weng, Yixuan
and Du, Yue
and Tian, Yuhang
and Zhou, Shoujun",
editor="Mao, Xian-Ling
and Ren, Zhaochun
and Yang, Muyun",
title="Overview of the NLPCC 2025 Shared Task 4: Multi-modal, Multilingual, and Multi-hop Medical Instructional Video Question Answering Challenge",
booktitle="Natural Language Processing and Chinese Computing",
year="2026",
publisher="Springer Nature Singapore",
address="Singapore",
pages="367--379",
isbn="978-981-95-3352-7"
}

@misc{m3med,
      title={M$^3$-Med: A Benchmark for Multi-lingual, Multi-modal, and Multi-hop Reasoning in Medical Instructional Video Understanding}, 
      author={Shenxi Liu and Kan Li and Mingyang Zhao and Yuhang Tian and Bin Li and Shoujun Zhou and Hongliang Li and Fuxia Yang},
      year={2025},
      eprint={2507.04289},
      archivePrefix={arXiv},
      primaryClass={cs.CV},
      url={https://arxiv.org/abs/2507.04289}, 
}

@article{scopingreview,
title = {Transforming Surgical Training With AI Techniques for Training, Assessment, and Evaluation: Scoping Review},
journal = {Journal of Medical Internet Research},
volume = {27},
year = {2025},
issn = {1438-8871},
doi = {https://doi.org/10.2196/58966},
url = {https://www.sciencedirect.com/science/article/pii/S1438887125014931},
author = {David Escobar-Castillejos and Ari Y Barrera-Animas and Julieta Noguez and Alejandra J Magana and Bedrich Benes}
}

@article{surgicalsimulation,
title = {Surgical Simulation: Virtual Reality to Artificial Intelligence},
journal = {Current Problems in Surgery},
volume = {61},
number = {11},
pages = {101625},
year = {2024},
issn = {0011-3840},
doi = {https://doi.org/10.1016/j.cpsurg.2024.101625},
url = {https://www.sciencedirect.com/science/article/pii/S0011384024001862},
author = {Elijah W. Riddle and Divya Kewalramani and Mayur Narayan and Daniel B. Jones}
}

@Article{Shahrezaei2024,
author={Shahrezaei, Aidin
and Sohani, Maryam
and Taherkhani, Soroush
and Zarghami, Seyed Yahya},
title={The impact of surgical simulation and training technologies on general surgery education},
journal={BMC Medical Education},
year={2024},
month={Nov},
day={13},
volume={24},
number={1},
pages={1297},
issn={1472-6920},
doi={10.1186/s12909-024-06299-w},
url={https://doi.org/10.1186/s12909-024-06299-w}
}

@article{onlinevideos,
title = {The Role of Online Videos in Teaching Procedural Skills in Postgraduate Medical Education: A Scoping Review},
journal = {Journal of Surgical Education},
volume = {79},
number = {5},
pages = {1295-1307},
year = {2022},
issn = {1931-7204},
doi = {https://doi.org/10.1016/j.jsurg.2022.05.009},
url = {https://www.sciencedirect.com/science/article/pii/S1931720422001118},
author = {Komal Srinivasa and Fiona Moir and Felicity Goodyear-Smith}
}

@article{onlinevideoteaching,
  title={How to Develop an Online Video for Teaching Health Procedural Skills: Tutorial for Health Educators New to Video Production},
  author={Komal Srinivasa and Amanda Charlton and Fiona Moir and Felicity Goodyear-Smith},
  journal={JMIR Medical Education},
  year={2023},
  volume={10},
  url={https://api.semanticscholar.org/CorpusID:270803777}
}

@article{makingeffective,
title = {Making Effective Educational Videos for Clinical Teaching},
journal = {Chest},
volume = {161},
number = {3},
pages = {764-772},
year = {2022},
issn = {0012-3692},
doi = {https://doi.org/10.1016/j.chest.2021.09.015},
url = {https://www.sciencedirect.com/science/article/pii/S0012369221039593},
author = {Ilana Roberts Krumm and Matthew C. Miles and Alison Clay and W. Graham {Carlos II} and Rosemary Adamson}
}

@InProceedings{temporal_answer_localization,
author="Cheng, Shuang
and Zhou, Zineng
and Liu, Jun
and Ye, Jian
and Luo, Haiyong
and Gu, Yang",
editor="Liu, Fei
and Duan, Nan
and Xu, Qingting
and Hong, Yu",
title="A Unified Framework for Optimizing Video Corpus Retrieval and Temporal Answer Grounding: Fine-Grained Modality Alignment and Local-Global Optimization",
booktitle="Natural Language Processing and Chinese Computing",
year="2023",
publisher="Springer Nature Switzerland",
address="Cham",
pages="199--210",
isbn="978-3-031-44699-3"
}

@inproceedings{video_corpus_retrieval, 
   series={SIGIR ’21},
   title={Video Corpus Moment Retrieval with Contrastive Learning},
   url={http://dx.doi.org/10.1145/3404835.3462874},
   DOI={10.1145/3404835.3462874},
   booktitle={Proceedings of the 44th International ACM SIGIR Conference on Research and Development in Information Retrieval},
   publisher={ACM},
   author={Zhang, Hao and Sun, Aixin and Jing, Wei and Nan, Guoshun and Zhen, Liangli and Zhou, Joey Tianyi and Goh, Rick Siow Mong},
   year={2021},
   month=July, pages={685–695},
   collection={SIGIR ’21} }

@Article{Youtube,
	author={Curran, Vernon
	and Simmons, Karla
	and Matthews, Lauren
	and Fleet, Lisa
	and Gustafson, Diana L.
	and Fairbridge, Nicholas A.
	and Xu, Xiaolin},
	title={YouTube as an Educational Resource in Medical Education: a Scoping Review},
	journal={Medical Science Educator},
	year={2020},
	month={Dec},
	day={01},
	volume={30},
	number={4},
	pages={1775-1782},
	issn={2156-8650},
	doi={10.1007/s40670-020-01016-w},
	url={https://doi.org/10.1007/s40670-020-01016-w}
}

@inproceedings{multishortcoming,
author = {Park, Jean and Jang, Kuk Jin and Alasaly, Basam and Mopidevi, Sriharsha and Zolensky, Andrew and Eaton, Eric and Lee, Insup and Johnson, Kevin},
title = {Assessing modality bias in video question answering benchmarks with multimodal large language models},
year = {2025},
isbn = {978-1-57735-897-8},
publisher = {AAAI Press},
url = {https://doi.org/10.1609/aaai.v39i19.34183},
doi = {10.1609/aaai.v39i19.34183},
booktitle = {Proceedings of the Thirty-Ninth AAAI Conference on Artificial Intelligence and Thirty-Seventh Conference on Innovative Applications of Artificial Intelligence and Fifteenth Symposium on Educational Advances in Artificial Intelligence},
articleno = {2210},
numpages = {9},
series = {AAAI'25/IAAI'25/EAAI'25}
}

@misc{videochallenges,
      title={Video Question Answering: Datasets, Algorithms and Challenges}, 
      author={Yaoyao Zhong and Junbin Xiao and Wei Ji and Yicong Li and Weihong Deng and Tat-Seng Chua},
      year={2022},
      eprint={2203.01225},
      archivePrefix={arXiv},
      primaryClass={cs.CV},
      url={https://arxiv.org/abs/2203.01225}, 
}

@Article{Tips,
author={Burgess, Annette
and van Diggele, Christie
and Roberts, Chris
and Mellis, Craig},
title={Tips for teaching procedural skills},
journal={BMC Medical Education},
year={2020},
month={Dec},
day={03},
volume={20},
number={2},
pages={458},
issn={1472-6920},
doi={10.1186/s12909-020-02284-1},
url={https://doi.org/10.1186/s12909-020-02284-1}
}

@article{useofvideos,
title = {Use of Videos by Health Care Professionals for Procedure Support in Acute Cardiac Care: A Scoping Review},
journal = {Heart, Lung and Circulation},
volume = {32},
number = {2},
pages = {143-155},
year = {2023},
issn = {1443-9506},
doi = {https://doi.org/10.1016/j.hlc.2022.10.004},
url = {https://www.sciencedirect.com/science/article/pii/S1443950622011337},
author = {Jacqueline Colgan and Sarah Kourouche and Geoffrey Tofler and Thomas Buckley}
}

@inproceedings{TimeCraft,
author = {Liu, Huabin and Ma, Xiao and Zhong, Cheng and Zhang, Yang and Lin, Weiyao},
title = {TimeCraft:\&nbsp;Navigate Weakly-Supervised Temporal Grounded Video Question Answering via\&nbsp;Bi-directional Reasoning},
year = {2024},
isbn = {978-3-031-72651-4},
publisher = {Springer-Verlag},
address = {Berlin, Heidelberg},
url = {https://doi.org/10.1007/978-3-031-72652-1_6},
doi = {10.1007/978-3-031-72652-1_6},
booktitle = {Computer Vision – ECCV 2024: 18th European Conference, Milan, Italy, September 29–October 4, 2024, Proceedings, Part V},
pages = {92–107},
numpages = {16},
location = {Milan, Italy}
}

@article{assessingmodality,
  title={Assessing Modality Bias in Video Question Answering Benchmarks with Multimodal Large Language Models},
  author={Jean Park and Kuk Jin Jang and Basam Alasaly and Sriharsha Mopidevi and Andrew Zolensky and Eric Eaton and Insup Lee and Kevin Johnson},
  journal={ArXiv},
  year={2024},
  volume={abs/2408.12763},
  url={https://api.semanticscholar.org/CorpusID:271947039}
}

@inproceedings{tvqa,
    title = "{TVQA}+: Spatio-Temporal Grounding for Video Question Answering",
    author = "Lei, Jie  and
      Yu, Licheng  and
      Berg, Tamara  and
      Bansal, Mohit",
    editor = "Jurafsky, Dan  and
      Chai, Joyce  and
      Schluter, Natalie  and
      Tetreault, Joel",
    booktitle = "Proceedings of the 58th Annual Meeting of the Association for Computational Linguistics",
    month = jul,
    year = "2020",
    address = "Online",
    publisher = "Association for Computational Linguistics",
    url = "https://aclanthology.org/2020.acl-main.730/",
    doi = "10.18653/v1/2020.acl-main.730",
    pages = "8211--8225"
}

@article{EffectiveEdu,
  title={Effective Educational Videos: Principles and Guidelines for Maximizing Student Learning from Video Content},
  author={Cynthia J. Brame},
  journal={CBE Life Sciences Education},
  year={2016},
  volume={15},
  url={https://api.semanticscholar.org/CorpusID:16260174}
}

@Article{Gupta2023,
author={Gupta, Deepak
and Attal, Kush
and Demner-Fushman, Dina},
title={A dataset for medical instructional video classification and question answering},
journal={Scientific Data},
year={2023},
month={Mar},
day={22},
volume={10},
number={1},
pages={158},
issn={2052-4463},
doi={10.1038/s41597-023-02036-y},
url={https://doi.org/10.1038/s41597-023-02036-y}
}

@misc{medcraft,
      title={Med-CRAFT: Automated Construction of Interpretable and Multi-Hop Video Workloads via Knowledge Graph Traversal}, 
      author={Shenxi Liu and Kan Li and Mingyang Zhao and Yuhang Tian and Shoujun Zhou and Bin Li},
      year={2025},
      eprint={2512.01045},
      archivePrefix={arXiv},
      primaryClass={cs.AI},
      url={https://arxiv.org/abs/2512.01045}, 
}

@misc{qwen35blog,
    title = {Qwen3.5: Accelerating Productivity with Native Multimodal Agents},
    url = {https://qwen.ai/blog?id=qwen3.5},
    author = {Qwen Team},
    month = {February},
    year = {2026}
}

@INPROCEEDINGS{mutualsl,
  author={Weng, Yixuan and Li, Bin},
  booktitle={ICASSP 2023 - 2023 IEEE International Conference on Acoustics, Speech and Signal Processing (ICASSP)}, 
  title={Visual Answer Localization with Cross-Modal Mutual Knowledge Transfer}, 
  year={2023},
  volume={},
  number={},
  pages={1-5},
  doi={10.1109/ICASSP49357.2023.10095026}}

@inproceedings{zhang2024improving,
	title={Improving Multilingual Temporal Answering Grounding in Single Video via LLM-Based Translation and OCR Enhancement},
	author={Zhang, Huan and Zheng, Chen and He, Yuanjing and Zhao, Yan and Lai, Yuxuan},
	booktitle={CCF International Conference on Natural Language Processing and Chinese Computing},
	pages={145--156},
	year={2024},
	organization={Springer}
}

@inproceedings{zhou2023improving,
	title={Improving Cross-Modal Visual Answer Localization in Chinese Medical Instructional Video Using Language Prompts},
	author={Zhou, Zineng and Liu, Jun and Cheng, Shuang and Luo, Haiyong and Gu, Yang and Ye, Jian},
	booktitle={CCF International Conference on Natural Language Processing and Chinese Computing},
	pages={221--232},
	year={2023},
	organization={Springer}
}

@inproceedings{li2023overview,
	title={Overview of the NLPCC 2023 shared task: Chinese medical instructional video question answering},
	author={Li, Bin and Weng, Yixuan and Guo, Hu and Sun, Bin and Li, Shutao and Luo, Yuhao and Qi, Mengyao and Liu, Xufei and Han, Yuwei and Liang, Haiwen and others},
	booktitle={CCF International Conference on Natural Language Processing and Chinese Computing},
	pages={233--242},
	year={2023},
	organization={Springer}
}

@inproceedings{yu2024mqua,
	title={MQuA: Multi-level Query-Video Augmentation for Multilingual Video Corpus Retrieval},
	author={Yu, Guyang and Bi, Xiaoyang and Tang, Jielong and Gu, Ming and Chen, Tianbai and Li, Zhiqiang and Zhu, Miankuan},
	booktitle={CCF International Conference on Natural Language Processing and Chinese Computing},
	pages={353--364},
	year={2024},
	organization={Springer}
}

@inproceedings{lei2023two,
	title={A two-stage Chinese medical video retrieval framework with LLM},
	author={Lei, Ningjie and Cai, Jinxiang and Qian, Yixin and Zheng, Zhilong and Han, Chao and Liu, Zhiyue and Huang, Qingbao},
	booktitle={CCF International Conference on Natural Language Processing and Chinese Computing},
	pages={211--220},
	year={2023},
	organization={Springer}
}

@inproceedings{ma2024multilingual,
	title={Multilingual Temporal Answer Grounding in Video Corpus with Enhanced Visual-Textual Integration},
	author={Ma, Tianxing and Hu, Yueyue and Jiang, Shuang and Yin, Zhenhao and Zang, Tianning},
	booktitle={CCF International Conference on Natural Language Processing and Chinese Computing},
	pages={471--483},
	year={2024},
	organization={Springer}
}

@inproceedings{cheng2023unified,
	title={A unified framework for optimizing video corpus retrieval and temporal answer grounding: fine-grained modality alignment and local-global optimization},
	author={Cheng, Shuang and Zhou, Zineng and Liu, Jun and Ye, Jian and Luo, Haiyong and Gu, Yang},
	booktitle={CCF International Conference on Natural Language Processing and Chinese Computing},
	pages={199--210},
	year={2023},
	organization={Springer}
}

@InProceedings{dima2025,
author="Wang, Yu
and Tan, Tianhao
and Wang, Yifei",
editor="Mao, Xian-Ling
and Ren, Zhaochun
and Yang, Muyun",
title="Hierarchical Indexing with Knowledge Enrichment for Multilingual Video Corpus Retrieval",
booktitle="Natural Language Processing and Chinese Computing",
year="2026",
publisher="Springer Nature Singapore",
address="Singapore",
pages="393--404",
isbn="978-981-95-3352-7"
}

@InProceedings{medecho2025,
author="Zhou, Yangchengyu
and Wu, Jiayuan
and Li, Yunze",
editor="Mao, Xian-Ling
and Ren, Zhaochun
and Yang, Muyun",
title="Multi-hop Knowledge-Enhanced Query Reasoning for Multi-modal Medical Video QA",
booktitle="Natural Language Processing and Chinese Computing",
year="2026",
publisher="Springer Nature Singapore",
address="Singapore",
pages="380--392",
isbn="978-981-95-3352-7"
}

@article{Li2022LearningTL,
  title={Learning To Locate Visual Answer In Video Corpus Using Question},
  author={Bin Li and Yixuan Weng and Bin Sun and Shutao Li},
  journal={ICASSP 2023 - 2023 IEEE International Conference on Acoustics, Speech and Signal Processing (ICASSP)},
  year={2022},
  pages={1-5},
  url={https://api.semanticscholar.org/CorpusID:252816090}
}

@misc{weng2022,
      title={Visual Answer Localization with Cross-modal Mutual Knowledge Transfer}, 
      author={Yixuan Weng and Bin Li},
      year={2022},
      eprint={2210.14823},
      archivePrefix={arXiv},
      primaryClass={cs.CV},
      url={https://arxiv.org/abs/2210.14823}, 
}

@ARTICLE{tovisual,
  author={Li, Shutao and Li, Bin and Sun, Bin and Weng, Yixuan},
  journal={IEEE Transactions on Pattern Analysis and Machine Intelligence}, 
  title={Towards Visual-Prompt Temporal Answer Grounding in Instructional Video}, 
  year={2024},
  volume={46},
  number={12},
  pages={8836-8853},
  doi={10.1109/TPAMI.2024.3411045}}

\end{document}